\documentclass[11pt,a4paper]{article}
\usepackage[hyperref]{acl2020}
\usepackage{times}
\usepackage{latexsym}

\usepackage{graphicx}
\usepackage{tabularx}
\usepackage{soul}
\usepackage{tabu}

\usepackage{epstopdf}
\usepackage[utf8]{inputenc}
\usepackage[T5,T1]{fontenc}

\usepackage{hyperref}
\usepackage{xstring}

\usepackage{color}

\usepackage{url}
\usepackage{booktabs}

\usepackage{multirow}
\usepackage{graphicx}
\usepackage{amsmath}
\usepackage{array}
\usepackage{xspace}

\makeatletter
\g@addto@macro{\UrlBreaks}{\UrlOrds}
\makeatother

\usepackage{microtype}

\aclfinalcopy %
\def\aclpaperid{1766} %

\newcommand{\elmooscar}{ELMo\textsubscript{OSCAR}\xspace}
\newcommand{\elmooscars}{ELMo\textsubscript{OSCAR}'s\xspace}
\newcommand{\elmowiki}{ELMo\textsubscript{Wikipedia}\xspace}
\newcommand{\elmowikis}{ELMo\textsubscript{Wikipedia}'s\xspace}
\newcommand{\elmooscarone}{ELMo\textsubscript{OSCAR(1)}\xspace}
\newcommand{\elmowikione}{ELMo\textsubscript{Wikipedia(1)}\xspace}
\newcommand{\elmooscarthree}{ELMo\textsubscript{OSCAR(3)}\xspace}
\newcommand{\elmowikithree}{ELMo\textsubscript{Wikipedia(3)}\xspace}
\newcommand{\elmooscarfive}{ELMo\textsubscript{OSCAR(5)}\xspace}
\newcommand{\elmowikifive}{ELMo\textsubscript{Wikipedia(5)}\xspace}
\newcommand{\elmooscarten}{ELMo\textsubscript{OSCAR(10)}\xspace}
\newcommand{\elmowikiten}{ELMo\textsubscript{Wikipedia(10)}\xspace}

\DeclareTextSymbolDefault{\OHORN}{T5}
\DeclareTextSymbolDefault{\UHORN}{T5}
\DeclareTextSymbolDefault{\ohorn}{T5}
\DeclareTextSymbolDefault{\uhorn}{T5}

\title{A Monolingual Approach to Contextualized Word Embeddings\\ for Mid-Resource Languages}

\author{Pedro Javier Ortiz Su\'arez$^{1,2}$ \qquad  Laurent Romary$^{1}$ \qquad Beno\^it Sagot$^{1}$\\
  $^{1}$Inria, Paris, France \\
  $^{2}$Sorbonne Universit\'e, Paris, France\\
  \texttt{\{pedro.ortiz, benoit.sagot, laurent.romary\}@inria.fr}\\}

\date{\today}

\begin{document}
\maketitle
\begin{abstract}
    We use the multilingual OSCAR corpus, extracted from Common Crawl via language classification, filtering and cleaning, to train monolingual contextualized word embeddings (ELMo) for five mid-resource languages. We then compare the performance of OSCAR-based and Wikipedia-based ELMo embeddings for these languages on the part-of-speech tagging and parsing tasks. We show that, despite the noise in the Common-Crawl-based OSCAR data, embeddings trained on OSCAR perform much better than monolingual embeddings trained on Wikipedia. They actually equal or improve the current state of the art in tagging and parsing for all five languages. In particular, they also improve over multilingual Wikipedia-based contextual embeddings (multilingual BERT), which almost always constitutes the previous state of the art, thereby showing that the benefit of a larger, more diverse corpus surpasses the cross-lingual benefit of multilingual embedding architectures.
\end{abstract}

\section{Introduction}
One of the key elements that has pushed the state of the art considerably in neural NLP in recent years has been the introduction and spread of transfer learning methods to the field. These methods can normally be classified in two categories according to how they are used:
\begin{itemize}
    \item \emph{Feature-based} methods, which involve pre-training real-valued vectors (``embeddings'') at the word, sentence, or paragraph level; and using them in conjunction with a specific architecture for each individual downstream task.
    \item \emph{Fine-tuning} methods, which introduce a minimal number of task-specific parameters, and instead copy the weights from a pre-trained network and then tune them to a particular downstream task.
\end{itemize}
Embeddings or language models can be divided into \emph{fixed}, meaning that they generate a single representation for each word in the vocabulary; and \emph{contextualized}, meaning that a representation is generated based on both the word and its surrounding context, so that a single word can have multiple representations, each one depending on how it is used.

In practice, most fixed embeddings are used as feature-based models. The most notable examples are \emph{word2vec} \citep{Mikolov:2013}, \emph{GloVe} \citep{Pennington:2014} and \emph{fastText} \citep{Mikolov:2018}. All of them are extensively used in a variety of applications nowadays. On the other hand, contextualized word representations and language models have been developed using both feature-based architectures, the most notable examples being ELMo and Flair \citep{Peters:2018,Akbik:2018}, and transformer based architectures, that are commonly used in a fine-tune setting, as is the case of GPT-1, GPT-2 \citep{Radford:2018,Radford:2019}, BERT and its derivatives \citep{Devlin:2018,Liu:2019,Lan:2019} and more recently T5 \citep{Raffel:2019}. All of them have repeatedly improved the state-of-the art in many downstream NLP tasks over the last year.

In general, the main advantage of using language models is that they are mostly built in an \emph{unsupervised} manner and they can be trained with raw, unannotated plain text. Their main drawback is that enormous quantities of data seem to be required to properly train them especially in the case of contextualized models, for which larger corpora are thought to be needed to properly address polysemy and cover the wide range of uses that commonly exist within languages.

For gathering data in a wide range of languages, Wikipedia is a commonly used option. It has been used to train fixed embeddings \citep{Al-rfou:2013,Bojanowski:2017} and more recently the multilingual BERT \citep{devlin2018mbert}, hereafter mBERT. However, for some languages, Wikipedia might not be large enough to train good quality contextualized word embeddings. Moreover, Wikipedia data all belong to the same specific genre and style. To address this problem, one can resort to crawled text from the internet; the largest and most widespread dataset of crawled text being Common Crawl.\footnote{\url{https://commoncrawl.org}} Such an approach generally solves the quantity and genre/style coverage problems but might introduce noise in the data, an issue which has earned the corpus some criticism, most notably by \citet{Trieu:2018} and \citet{Radford:2019}. Using Common Crawl also leads to data management challenges as the corpus is distributed in the form of a large set of plain text each containing a large quantity of unclassified multilingual documents from different websites.

In this paper we study the trade-off between quantity and quality of data for training contextualized representations. To this end, we use the OSCAR corpus \citep{ortiz2019asynchronous}, a freely available\footnote{\url{https://oscar-corpus.com}} multilingual dataset obtained by performing language classification, filtering and cleaning of the whole Common Crawl corpus.\footnote{\href{http://commoncrawl.org/2018/11/november-2018-crawl-archive-now-available/}{Snapshot from November 2018}} OSCAR was created following the approach of \citet{Grave:2018} but proposing a simple improvement on their filtering method. We then train OSCAR-based and Wikipedia-based ELMo contextualized word embeddings \citep{Peters:2018} for 5 languages: Bulgarian, Catalan, Danish, Finnish and Indonesian. We evaluate the models by attaching them to the to UDPipe 2.0 architecture \citep{Straka:2018,Straka:2019} for dependency parsing and part-of-speech (POS) tagging. We show that the models using the OSCAR-based ELMo embeddings consistently outperform the Wikipedia-based ones, suggesting that big high-coverage noisy corpora might be better than small high-quality narrow-coverage corpora for training contextualized language representations\footnote{Both the Wikipedia- and the OSCAR-based embeddings for these 5 languages are available at: \url{https://oscar-corpus.com/\#models}.}. We also establish a new state of the art for both POS tagging and dependency parsing in 6 different treebanks covering all 5 languages.

The  structure  of  the  paper  is  as  follows. In Section 2 we describe the recent related work. In Section 3  we present, compare and analyze the corpora used to train our contextualized embeddings, and the treebanks used to train our POS tagging and parsing models. In Section 4 we examine and describe in detail the model used for our contextualized word representations, as well as the parser and the tagger we chose to evaluate the impact of corpora in the embeddings' performance in downstream tasks. Finally we provide an analysis of our results in Section 5 and in Section 6 we present our conclusions.

\section{Related work}

Since the introduction of \emph{word2vec} \citep{Mikolov:2013}, many attempts have been made to create multilingual language representations; for fixed word embeddings the most remarkable works are those of \citep{Al-rfou:2013} and \citep{Bojanowski:2017} who created word embeddings for a large quantity of languages using Wikipedia, and later \citep{Grave:2018} who trained the fastText word embeddings for 157 languages using Common Crawl and who in fact showed that using crawled data significantly increased the performance of the embeddings especially for mid- to low-resource languages.

Regarding contextualized models, the most notable non-English contribution has been that of the mBERT \citep{devlin2018mbert}, which is distributed as (i)~a single multilingual model for 100 different languages trained on Wikipedia data, and as (ii)~a single multilingual model for both Simplified and Traditional Chinese. Four monolingual fully trained ELMo models have been distributed for Japanese, Portuguese, German and Basque\footnote{\url{https://allennlp.org/elmo}}; 44 monolingual ELMo models\footnote{\url{https://github.com/HIT-SCIR/ELMoForManyLangs}} where also released by the \emph{HIT-SCIR} team \citep{Che:2018} during the \emph{CoNLL 2018 Shared Task} \citep{Zeman:2018}, but their training sets where capped at 20 million words. A German BERT \citep{chan2019german} as well as a French BERT model (called CamemBERT) \citep{Martin:2019} have also been released. In general no particular effort in creating a set of high-quality monolingual contextualized representations has been shown yet, or at least not on a scale that is comparable with what was done for fixed word embeddings.

For dependency parsing and POS tagging the most notable non-English specific contribution is that of the \emph{CoNLL 2018 Shared Task} \citep{Zeman:2018}, where the 1\textsuperscript{st} place (LAS Ranking) was awarded to the \emph{HIT-SCIR} team \citep{Che:2018} who used \citet{Dozat:2017b}'s \emph{Deep Bi-affine parser} and its extension described in \citep{Dozat:2017}, coupled with deep contextualized ELMo embeddings \citep{Peters:2018} (capping the training set at 20 million words). The 1\textsuperscript{st} place in universal POS tagging was awarded to \citet{Smith:2018} who used two separate instances of \citet{Bohnet:2018}'s tagger.

More recent developments in POS tagging and parsing include those of \citet{Straka:2019} which couples another CoNLL 2018 shared task participant, UDPipe 2.0 \citep{Straka:2018}, with mBERT greatly improving the scores of the original model, and UDify \citep{Kondratyuk:2019}, which adds an extra attention layer on top of mBERT plus a Deep Bi-affine attention layer for dependency parsing and a Softmax layer for POS tagging. UDify is actually trained by concatenating the training sets of 124 different UD treebanks, creating a single POS tagging and dependency parsing model that works across 75 different languages.

\section{Corpora}
We train ELMo contextualized word embeddings for 5 languages: Bulgarian, Catalan, Danish, Finnish and Indonesian. We train one set of embeddings using only Wikipedia data, and another set using only  Common-Crawl-based OSCAR data. We chose these languages primarily because they are morphologically and typologically different from one another, but also because all of the OSCAR datasets for these languages were of a sufficiently manageable size such that the ELMo pre-training was doable in less than one month. Contrary to \emph{HIT-SCIR} team \citep{Che:2018}, we do not impose any cap on the amount of data, and instead use the entirety of Wikipedia or OSCAR for each of our 5 chosen languages.

\subsection{Wikipedia}

\begin{table}[t!]
    \centering\small
    \scalebox{0.91}{
        \begin{tabular}{lrrrr}\toprule
            Language   & \multicolumn{1}{l}{Size} & \multicolumn{1}{l}{\#Ktokens} & \multicolumn{1}{l}{\#Kwords} & \multicolumn{1}{l}{\#Ksentences} \\ \midrule
            Bulgarian  & 609M                     & 64,190                        & 54,748                       & 3,685                            \\
            Catalan    & 1.1G                     & 211,627                       & 179,108                      & 8,293                            \\
            Danish     & 338M                     & 60,644                        & 52,538                       & 3,226                            \\
            Finnish    & 669M                     & 89,580                        & 76,035                       & 6,847                            \\
            Indonesian & 488M                     & 80,809                        & 68,955                       & 4,298                            \\
            \bottomrule
        \end{tabular}
    }
    \caption{Size of Wikipedia corpora, measured in bytes, thousands of tokens, words and sentences.}
    \label{tab:Wikipedia}
\end{table}

Wikipedia is the biggest online multilingual open encyclopedia, comprising more than 40 million articles in 301 different languages. Because articles are curated by language and written in an open collaboration model, its text tends to be of very high-quality in comparison to other free online resources. This is why Wikipedia has been extensively used in various NLP applications \citep{Wu:2010,Mihalcea:2007,Al-rfou:2013,Bojanowski:2017}. We downloaded the XML Wikipedia dumps\footnote{XML dumps from April 4, 2019.} and extracted the plain-text from them using the \texttt{wikiextractor.py} script\footnote{Available \href{https://github.com/attardi/wikiextractor}{here}.} from Giuseppe Attardi. We present the number of words and tokens available for each of our 5 languages in Table \ref{tab:Wikipedia}. We decided against deduplicating the Wikipedia data as the corpora are already quite small. We tokenize the 5 corpora using \emph{UDPipe} \citep{Straka:2017}.

\subsection{OSCAR}

Common Crawl is a non-profit organization that produces and maintains an open, freely available repository of crawled data from the web. Common Crawl's complete archive consists of petabytes of monthly snapshots collected since 2011. {} Common Crawl snapshots are not classified by language, and contain a certain level of noise (e.g.~one-word ``sentences'' such as ``OK'' and ``Cancel'' are unsurprisingly very frequent).

This is what motivated the creation of the freely available multilingual OSCAR corpus \citep{ortiz2019asynchronous}, extracted from the November 2018 snapshot, which amounts to more than 20 terabytes of plain-text. In order to create OSCAR from this Common Crawl snapshot, \citet{ortiz2019asynchronous}  reproduced the pipeline proposed by \citep{Grave:2018} to process, filter and classify Common Crawl. More precisely,  language classification was performed using the \emph{fastText} linear classifier \citep{Joulin:2016,Joulin:2017}, which was trained by \citet{Grave:2018} to recognize 176 languages and was shown to have an extremely good accuracy to processing time trade-off. The filtering step as performed by \citet{Grave:2018} consisted in only keeping the lines exceeding 100 bytes in length.\footnote{Script available \href{https://github.com/facebookresearch/fastText/blob/master/crawl/process_wet_file.sh}{here}.} However, considering that Common Crawl is a mutilingual UTF-8 encoded corpus, this 100-byte threshold creates a huge disparity between ASCII and non-ASCII encoded languages. The filtering step used to create OSCAR therefore consisted in only keeping the lines containing at least 100 UTF-8-encoded characters. Finally, as in  \citep{Grave:2018}, the OSCAR corpus is deduplicated, i.e.~for each language, only one occurrence of a given line is included.

As we did for Wikipedia, we tokenize OSCAR corpora for the 5 languages we chose for our study using UDPipe. Table \ref{tab:CC} provides quantitative information about the 5 resulting tokenized corpora.

We note that the original Common-Crawl-based corpus created by \citet{Grave:2018} to train fastText is not freely available. Since running the experiments described in this paper, a new architecture for creating a Common-Crawl-based corpus named CCNet \citep{wenzek2019ccnet} has been published, although it includes specialized filtering which might result in a cleaner corpus compared to OSCAR, the resulting CCNet corpus itself was not published. Thus we chose to keep OSCAR as it remains the only very large scale, Common-Crawl-based corpus currently available and easily downloadable.

\begin{table}[t]
    \centering\small
    \scalebox{0.91}{
        \begin{tabular}{lrrrr}\toprule
            Language   & \multicolumn{1}{l}{Size} & \multicolumn{1}{l}{\#Ktokens} & \multicolumn{1}{l}{\#Kwords} & \multicolumn{1}{l}{\#Ksentences} \\ \midrule
            Bulgarian  & 14G                      & 1,466,051                     & 1,268,115                    & 82,532                           \\
            Catalan    & 4.3G                     & 831,039                       & 729,333                      & 31,732                           \\
            Danish     & 9.7G                     & 1,828,881                     & 1,620,091                    & 99,766                           \\
            Finnish    & 14G                      & 1,854,440                     & 1,597,856                    & 142,215                          \\
            Indonesian & 16G                      & 2,701,627                     & 2,394,958                    & 140,138                          \\
            \bottomrule
        \end{tabular}
    }
    \caption{Size of OSCAR subcorpora, measured in bytes, thousands of tokens, words and sentences.}
    \label{tab:CC}
\end{table}

\subsection{Noisiness}

We wanted to address \citep{Trieu:2018} and \citep{Radford:2019}'s criticisms of Common Crawl, so we devised a simple method to measure how noisy the OSCAR corpora were for our 5 languages. We randomly extract a number of lines from each corpus, such that the resulting random sample contains one million words.\footnote{We remove tokens that are capitalized or contain less than 4 UTF-8 encoded characters, allowing us to remove bias against Wikipedia, which traditionally contains a large quantity of proper nouns and acronyms.} We test if the words are in the corresponding \emph{GNU Aspell}\footnote{\url{http://aspell.net/}} dictionary. We repeat this task for each of the 5 languages, for both the OSCAR and the Wikipedia corpora. We compile in Table \ref{tab:OOV} the number of out-of-vocabulary tokens for each corpora.

As expected, this simple metric shows that in general the OSCAR samples contain more out-of-vocabulary words than the Wikipedia ones. However the difference in magnitude between the two is strikingly lower that one would have expected in view of the criticisms by \citet{Trieu:2018} and \citet{Radford:2019}, thereby validating the usability of Common Crawl data when it is properly filtered, as was achieved by the OSCAR creators. We even observe that, for Danish, the number of out-of-vocabulary words in OSCAR is lower than that in Wikipedia.

\begin{table}[t]
    \centering\small
    \begin{tabular}{lrr}\toprule
        Language   & \multicolumn{1}{l}{OOV Wikipedia} & \multicolumn{1}{l}{OOV OSCAR} \\ \midrule
        Bulgarian  & 60,879                            & 66,558                        \\
        Catalan    & 34,919                            & 79,678                        \\
        Danish     & 134,677                           & 123,299                       \\
        Finnish    & 266,450                           & 267,525                       \\
        Indonesian & 116,714                           & 124,607                       \\
        \bottomrule
    \end{tabular}
    \caption{Number of out-of-vocabulary words in random samples of 1M words for OSCAR and Wikipedia.}
    \label{tab:OOV}
\end{table}

\section{Experimental Setting}

The main goal of this paper is to show the impact of training data on contextualized word representations when applied in particular downstream tasks. To this end, we train different versions of the \emph{Embeddings from Language Models} (ELMo) \citep{Peters:2018} for both the Wikipedia and OSCAR corpora, for each of our selected 5 languages. We save the models' weights at different number of epochs for each language, in order to test how corpus size affect the embeddings and to see whether and when overfitting happens when training elmo on smaller corpora.

We take each of the trained ELMo models and use them in conjunction with the UDPipe 2.0 \citep{Straka:2018,Straka:2019} architecture for dependency parsing and POS-tagging to test our models. We train UDPipe 2.0 using gold tokenization and segmentation for each of our ELMo models, the only thing that changes from training to training is the ELMo model as hyperparameters always remain at the default values (except for number of training tokens) \citep{Peters:2018}.

\subsection{Contextualized word embeddings}

\emph{Embeddings from Language Models} (ELMo) \citep{Peters:2018} is an LSTM-based language model. More precisely, it uses a bidirectional language model, which combines a forward and a backward LSTM-based language model. ELMo also computes a context-independent token representation via a CNN over characters.

We train ELMo models for Bulgarian, Catalan, Danish, Finnish and Indonesian using the OSCAR corpora on the one hand and the Wikipedia corpora on the other. We train each model for 10 epochs, as was done for the original English ELMo \citep{Peters:2018}. We save checkpoints at 1\textsuperscript{st}, 3\textsuperscript{rd} and 5\textsuperscript{th} epoch in order to investigate some concerns about possible overfitting for smaller corpora (Wikipedia in this case) raised by the original ELMo authors.\footnote{\url{https://github.com/allenai/bilm-tf/issues/135}}

\subsection{UDPipe 2.0} \label{udpipe-future}

For our POS tagging and dependency parsing evaluation, we use UDPipe 2.0, which has a freely available and ready to use implementation.\footnote{\url{https://github.com/CoNLL-UD-2018/UDPipe-Future}} This architecture was submitted as a participant to the \emph{2018 CoNLL Shared Task} \citep{Zeman:2018}, obtaining the 3\textsuperscript{rd} place in LAS ranking. UDPipe 2.0 is a multi-task model that predicts POS tags, lemmas and dependency trees jointly.

The original UDPipe 2.0 implementation calculates 3 different embeddings, namely:

\begin{itemize}
    \item \emph{Pre-trained word embeddings}: In the original implementation, the Wikipedia version of fastText embeddings is used \citep{Bojanowski:2017}; we replace them in favor of the newer Common-Crawl-based fastText embeddings trained by \citet{Grave:2018}.
    \item \emph{Trained word embeddings}: Randomly initialized word representations that are trained with the rest of the network.
    \item \emph{Character-level word embeddings}: Computed using bi-directional GRUs of dimension 256. They represent every UTF-8 encoded character with two 256 dimensional vectors, one for the forward and one for the backward layer. This two vector representations are concatenated and are trained along the whole network.
\end{itemize}

After the CoNLL 2018 Shared Task, the UDPipe 2.0 authors added the option to concatenate contextualized representations to the embedding section of the network \citep{Straka:2019}, we use this new implementation and we concatenate our pretrained deep contextualized ELMo embeddings to the three embeddings mentioned above.

Once the embedding step is completed, the concatenation of all vector representations for a word are fed to two shared bidirectional LSTM \citep{Hochreiter:1997} layers. The output of these two BiLSTMS is then fed to two separate specific LSTMs:
\begin{itemize}
    \item The tagger- and lemmatizer-specific bidirectional LSTMs, with Softmax classifiers on top, which process its output and generate UPOS, XPOS, UFeats and Lemmas. The lemma classifier also takes the character-level word embeddings as input.

    \item The parser-specific bidirectional LSTM layer, whose output is then passed to a bi-affine attention layer \citep{Dozat:2017b} producing labeled dependency trees.
\end{itemize}

\subsection{Treebanks}

\begin{table}[t!]
    \centering\small
    \begin{tabular}{lrr}\toprule
        Treebank       & \multicolumn{1}{l}{\#Ktokens} & \multicolumn{1}{l}{\#Ksentences} \\ \midrule
        Bulgarian-BTB  & 156                           & 11                               \\
        Catalan-AnCora & 530                           & 17                               \\
        Danish-DDT     & 100                           & 6                                \\
        Finnish-FTB    & 159                           & 19                               \\
        Finnish-TDT    & 202                           & 15                               \\
        Indonesian-GSD & 121                           & 6                                \\
        \bottomrule
    \end{tabular}
    \caption{Size of treebanks, measured in thousands of tokens and sentences.}
    \label{tab:treeb}
\end{table}

To train the selected parser and tagger (cf.~Section \ref{udpipe-future}) and evaluate the pre-trained language models in our 5 languages, we run our experiments using the Universal Dependencies (UD)\footnote{\url{https://universaldependencies.org}} paradigm and its corresponding UD POS tag set \citep{petrov2011universal}. We use all the treebanks available for our five languages in the UD treebank collection version 2.2 \citep{ud22}, which was used for the CoNLL 2018 shared task, thus we perform our evaluation tasks in 6 different treebanks (see Table~\ref{tab:treeb} for treebank size information).
\begin{itemize}
    \item \emph{Bulgarian BTB}: Created at the Institute of Information and Communication Technologies, Bulgarian Academy of Sciences, it consists of legal documents, news articles and fiction pieces.
    \item \emph{Catalan-AnCora}: Built on top of the Spanish-Catalan \emph{AnCora corpus} \citep{Taule:2008}, it contains mainly news articles.
    \item \emph{Danish-DDT}: Converted from the \emph{Danish Dependency Treebank} \citep{Buch:2003}. It includes news articles, fiction and non fiction texts and oral transcriptions.
    \item \emph{Finnish-FTB}: Consists of manually annotated grammatical examples from VISK\footnote{\url{http://scripta.kotus.fi/visk}} (The Web Version of the Large Grammar of Finnish).
    \item \emph{Finnish-TDT}: Based on the Turku Dependency Treebank (TDT). Contains texts from Wikipedia, Wikinews, news articles, blog entries, magazine articles, grammar examples, Europarl speeches, legal texts and fiction.
    \item \emph{Indonesian-GSD}: Includes mainly blog entries and news articles.
\end{itemize}

\section{Results \& Discussion}

\begin{table}[ht!]
    \centering\small
    \scalebox{0.95}{
        \begin{tabular}{@{}llccc@{}}\toprule
            Treebank       & Model       & UPOS              & UAS               & LAS               \\
            \midrule
                           & UDify       & 98.89             & 95.54             & 92.40             \\
                           & UDPipe 2.0  & 98.98             & 93.38             & 90.35             \\
            Bulgarian BTB  & +mBERT      & \underline{99.20} & \underline{95.34} & \underline{92.62} \\
                           & +\elmowiki  & 99.17             & 94.93             & 92.05             \\
                           & +\elmooscar & \textbf{99.40}    & \textbf{96.01}    & \textbf{93.56}    \\
            \midrule
                           & UDify       & 98.89             & \underline{94.25} & 92.33             \\
                           & UDPipe 2.0  & 98.88             & 93.22             & 91.06             \\
            Catalan-AnCora & +mBERT      & \textbf{99.06}    & \textbf{94.49}    & \underline{92.74} \\
                           & +\elmowiki  & \underline{99.05} & 93.99             & 92.24             \\
                           & +\elmooscar & \textbf{99.06}    & \textbf{94.49}    & \textbf{92.88}    \\
            \midrule
                           & UDify       & 97.50             & 87.76             & 84.50             \\
                           & UDPipe 2.0  & 97.78             & 86.88             & 84.31             \\
            Danish-DDT     & +mBERT      & 98.21             & \underline{89.32} & \underline{87.24} \\
                           & +\elmowiki  & \underline{98.45} & 89.05             & 86.92             \\
                           & +\elmooscar & \textbf{98.62}    & \textbf{89.84}    & \textbf{87.95}    \\
            \midrule
                           & UDify       & 93.80             & 86.37             & 81.40             \\
                           & UDPipe 2.0  & 96.65             & 90.68             & 87.89             \\
            Finnish-FTB    & +mBERT      & 96.97             & 91.68             & 89.02             \\
                           & +\elmowiki  & \underline{97.27} & \underline{92.05} & \underline{89.62} \\
                           & +\elmooscar & \textbf{98.13}    & \textbf{93.81}    & \textbf{92.02}    \\
            \midrule
                           & UDify       & 94.43             & 86.42             & 82.03             \\
                           & UDPipe 2.0  & 97.45             & 89.88             & 87.46             \\
            Finnish-TDT    & +mBERT      & 97.57             & \underline{91.66} & \underline{89.49} \\
                           & +\elmowiki  & \underline{97.65} & 91.60             & 89.34             \\
                           & +\elmooscar & \textbf{98.36}    & \textbf{93.54}    & \textbf{91.77}    \\
            \midrule
                           & UDify       & 93.36             & 86.45             & 80.10             \\
                           & UDPipe 2.0  & 93.69             & 85.31             & 78.99             \\
            Indonesian-GSD & +mBERT      & \underline{94.09} & \underline{86.47} & \underline{80.40} \\
                           & +\elmowiki  & 93.94             & 86.16             & 80.10             \\
                           & +\elmooscar & \textbf{94.12}    & \textbf{86.49}    & \textbf{80.59}    \\
            \bottomrule
        \end{tabular}
    }
    \caption{Scores from \mbox{UDPipe~2.0} \protect\citep[from][]{Kondratyuk:2019}, the previous state-of-the-art models UDPipe 2.0+mBERT \protect\citep{Straka:2019} and UDify \protect\citep{Kondratyuk:2019}, and our ELMo-enhanced UDPipe 2.0 models. Test scores are given for UPOS, UAS and LAS in all five languages. Best scores are shown in bold, second best scores are underlined.}
    \label{tab:parse}
\end{table}

\begin{table*}[htbp]
    \centering\small
    \scalebox{0.91}{
        \begin{tabular}{@{}llccc@{}}\toprule
            Treebank       & Model            & UPOS              & UAS               & LAS               \\ \midrule
                           & UDPipe 2.0       & 98.98             & 93.38             & 90.35             \\
                           & +\elmowikione    & 98.81             & 93.60             & 90.21             \\
                           & +\elmowikithree  & 99.01             & 94.32             & 91.36             \\
                           & +\elmowikifive   & 99.03             & 94.32             & 91.38             \\
            Bulgarian BTB  & +\elmowikiten    & \underline{99.17} & \underline{94.93} & \underline{92.05} \\
                           & +\elmooscarone   & 99.28             & 95.45             & 92.98             \\
                           & +\elmooscarthree & 99.34             & 95.58             & 93.12             \\
                           & +\elmooscarfive  & 99.34             & 95.63             & 93.25             \\
                           & +\elmooscarten   & \textbf{99.40}    & \textbf{96.01}    & \textbf{93.56}    \\
            \midrule
                           & UDPipe 2.0       & 98.88             & 93.22             & 91.06             \\
                           & +\elmowikione    & 98.93             & 93.24             & 91.21             \\
                           & +\elmowikithree  & 99.02             & 93.75             & 91.93             \\
                           & +\elmowikifive   & 99.04             & 93.86             & 92.05             \\
            Catalan-AnCora & +\elmowikiten    & \underline{99.05} & \underline{93.99} & \underline{92.24} \\
                           & +\elmooscarone   & 99.07             & 93.92             & 92.29             \\
                           & +\elmooscarthree & \textbf{99.10}    & 94.29             & 92.69             \\
                           & +\elmooscarfive  & 99.07             & 94.38             & 92.75             \\
                           & +\elmooscarten   & 99.06             & \textbf{94.49}    & \textbf{92.88}    \\
            \midrule
                           & UDPipe 2.0       & 97.78             & 86.88             & 84.31             \\
                           & +\elmowikione    & 97.47             & 86.98             & 84.15             \\
                           & +\elmowikithree  & 98.03             & 88.16             & 85.81             \\
                           & +\elmowikifive   & 98.15             & 88.24             & 85.96             \\
            Danish-DDT     & +\elmowikiten    & \underline{98.45} & \underline{89.05} & \underline{86.92} \\
                           & +\elmooscarone   & 98.50             & 89.47             & 87.43             \\
                           & +\elmooscarthree & 98.59             & 89.68             & 87.77             \\
                           & +\elmooscarfive  & 98.59             & 89.46             & 87.64             \\
                           & +\elmooscarten   & \textbf{98.62}    & \textbf{89.84}    & \textbf{87.95}    \\
            \bottomrule
        \end{tabular}
    }
    ~
    \scalebox{0.91}{
        \begin{tabular}{@{}llccc@{}}\toprule
            Treebank       & Model            & UPOS              & UAS               & LAS               \\
            \midrule
                           & UDPipe 2.0       & 96.65             & 90.68             & 87.89             \\
                           & +\elmowikione    & 95.86             & 89.63             & 86.39             \\
                           & +\elmowikithree  & 96.76             & 91.02             & 88.27             \\
                           & +\elmowikifive   & 96.97             & 91.66             & 89.04             \\
            Finnish-FTB    & +\elmowikiten    & \underline{97.27} & \underline{92.05} & \underline{89.62} \\
                           & +\elmooscarone   & 97.91             & 93.41             & 91.43             \\
                           & +\elmooscarthree & 98.00             & \textbf{93.99}    & 91.98             \\
                           & +\elmooscarfive  & \textbf{98.15}    & 93.98             & \textbf{92.24}    \\
                           & +\elmooscarten   & 98.13             & 93.81             & 92.02             \\
            \midrule
                           & UDPipe 2.0       & 97.45             & 89.88             & 87.46             \\
                           & +\elmowikione    & 96.73             & 89.11             & 86.33             \\
                           & +\elmowikithree  & 97.55             & 90.84             & 88.50             \\
                           & +\elmowikifive   & 97.55             & 91.11             & 88.88             \\
            Finnish-TDT    & +\elmowikiten    & \underline{97.65} & \underline{91.60} & \underline{89.34} \\
                           & +\elmooscarone   & 98.27             & 93.03             & 91.29             \\
                           & +\elmooscarthree & 98.38             & \textbf{93.60}    & \textbf{91.83}    \\
                           & +\elmooscarfive  & \textbf{98.39}    & 93.57             & 91.80             \\
                           & +\elmooscarten   & 98.36             & 93.54             & 91.77             \\
            \midrule
                           & UDPipe 2.0       & 93.69             & 85.31             & 78.99             \\
                           & +\elmowikione    & 93.70             & 85.81             & 79.46             \\
                           & +\elmowikithree  & 93.90             & 86.04             & 79.72             \\
                           & +\elmowikifive   & 94.04             & 85.93             & 79.97             \\
            Indonesian-GSD & +\elmowikiten    & \underline{93.94} & \underline{86.16} & \underline{80.10} \\
                           & +\elmooscarone   & 93.95             & 86.25             & 80.23             \\
                           & +\elmooscarthree & 94.00             & 86.21             & 80.14             \\
                           & +\elmooscarfive  & \textbf{94.23}    & 86.37             & 80.40             \\
                           & +\elmooscarten   & 94.12             & \textbf{86.49}    & \textbf{80.59}    \\
            \bottomrule
        \end{tabular}
    }
    \caption{UPOS, UAS and LAS scores for the UDPipe 2.0 baseline reported by \protect\citep{Kondratyuk:2019}, plus the scores for checkpoints at 1, 3, 5 and 10 epochs for all the \elmooscar and \elmowiki. All scores are test scores. Best \elmooscar scores are shown in bold while best \elmowiki scores are underlined.}
    \label{tab:ablation}
\end{table*}

\subsection{Parsing and POS tagging results}
We use UDPipe 2.0 without contextualized embeddings as our baseline for POS tagging and dependency parsing. However, we did not train the model without contextualized word embedding ourselves. We instead take the scores as they are reported in \citep{Kondratyuk:2019}. We also compare our UDPipe 2.0 + ELMo models against the state-of-the-art results (assuming gold tokenization) for these languages, which are either UDify \citep{Kondratyuk:2019} or UDPipe 2.0 + mBERT \citep{Straka:2019}.

Results for UPOS, UAS and LAS are shown in Table \ref{tab:parse}. We obtain the state of the art for the three metrics in each of the languages with the UDPipe 2.0 + \elmooscar models. We also see that in every single case the UDPipe 2.0 + \elmooscar result surpasses the UDPipe 2.0 + \elmowiki one, suggesting that the size of the pre-training data plays an important role in downstream task results. This is also supports our hypothesis that the OSCAR corpora, being multi-domain, exhibits a better coverage of the different styles, genres and uses present at least in these 5 languages.

Taking a closer look at the results for Danish, we see that \elmowiki, which was trained with a mere 300MB corpus, does not show any sign of overfitting, as the UDPipe 2.0 + \elmowiki results considerably improve the UDPipe 2.0 baseline. This is the case for all of our \elmowiki models as we never see any evidence of a negative impact when we add them to the baseline model. In fact, the results of UDPipe 2.0 + \elmowiki give better than previous state-of-the-art results in all metrics for the Finnish-FTB and in UPOS for the Finnish-TDT. The results for Finnish are actually quite interesting, as mBERT was pre-trained on Wikipedia and here we see that the multilingual setting in which UDify was fine-tuned exhibits sub-baseline results for all metrics, and that the UDPipe + mBERT scores are often lower than those of our UDPipe 2.0 + \elmowiki. This actually suggests that even though the multilingual approach of mBERT (in pre-training) or UDify (in pre-training and fine-tuning) leads to better performance for  high-resource languages or languages that are closely related to high-resource languages, it might also significantly degrade the representations for more isolated or even simply more morphologically rich languages like Finnish. In contrast, our monolingual approach with UDPipe 2.0 + \elmooscar improves the previous SOTA considerably, by more than 2 points for some metrics. Note however that Indonesian, which might also be seen as a relatively isolated language, does not behave in the same way as Finnish.

\subsection{Impact of the number of training epochs}

An important topic we wanted to address with our experiments was that of \emph{overfitting} and the number of epochs one should train the contextualized embeddings for. The ELMo authors have expressed that increasing the number of training epochs is generally better, as they argue that training the ELMo model for longer reduces held-out perplexity and further improves downstream task performance.\footnote{Their comments on the matter can be found \href{https://github.com/allenai/bilm-tf/issues/135}{here}.} This is why we intentionally fully pre-trained the \elmowiki to the 10 epochs of the original ELMo paper, as its authors also expressed concern over the possibility of overfitting for smaller corpora. We thus save checkpoints for each of our ELMo model at the 1, 3, 5 and 10 epoch marks so that we can properly probe for overfitting. The scores of all checkpoints are reported in Table \ref{tab:ablation}. Here again we do not train the UDPipe 2.0 baselines without embedding, we just report the scores published in \citet{Kondratyuk:2019}.

The first striking finding is that even though all our Wikipedia data sets are smaller than 1GB in size (except for Catalan), none of the \elmowiki models show any sign of overfitting, as the results continue to improve for all metrics the more we train the ELMo models, with the best results consistently being those of the fully trained 10 epoch ELMos. For all of our Wikipedia models, but those of Catalan and Indonesian, we see sub-baseline results at 1 epoch; training the model for longer is better, even if the corpora are small in size.

\elmooscar models exhibit exactly the same behavior as \elmowiki models where the scores continue to improve the longer they are pre-trained, except for the case of Finnish. Here we actually see an unexpected behavior where the model performance caps around the 3\textsuperscript{rd} to 5\textsuperscript{th} epoch. This is surprising because the Finnish OSCAR corpus is more than 20 times bigger than our smallest Wikipedia corpus, the Danish Wikipedia, that did not exhibit this behavior. As previously mentioned Finnish is morphologically richer than the other languages in which we trained ELMo, we hypothesize that the representation space given by the ELMo embeddings might not be sufficiently big to extract more features from the Finnish OSCAR corpus beyond the 5\textsuperscript{th} epoch mark, however in order to test this we would need to train a larger language model like BERT which is sadly beyond our computing infrastructure limits (cf. Subsection \ref{cost}). However we do note that pre-training our current language model architectures in a morphologically rich language like Finnish might actually better expose the limits of our existing approaches to language modeling.

One last thing that it is important to note with respect to the number of training epochs is that even though we fully pre-trained our \elmowikis and \elmooscars to the recommended 10 epoch mark, and then compared them against one another, the number of training steps between both pre-trained models differs drastically due to the big difference in corpus size (for Indonesian, for instance, 10 epochs correspond to 78K steps for \elmowiki and to 2.6M steps for OSCAR; the complete picture is provided in the Appendix, in Table~\ref{tab:steps}). In fact, we can see in Table \ref{tab:ablation} that all the UDPipe 2.0 + \elmooscarone perform better than the UDPipe 2.0 + \elmowikione models across all metrics. Thus we believe that talking in terms of training steps as opposed to training epochs might be a more transparent way of comparing two pre-trained models.

\subsection{Computational cost and carbon footprint}\label{cost}

Considering the discussion above, we believe an interesting follow-up to our experiments would be training the ELMo models for more of the languages included in the OSCAR corpus. However training ELMo is computationally costly, and one way to estimate this cost, as pointed out by \citet{strubell:2019}, is by using the training times of each model to compute both power consumption and CO\textsubscript{2} emissions.

\begin{table}[t]
    \centering\small
    \scalebox{0.93}{
        \begin{tabular}{@{}lrrrrr@{}}\toprule
            Language                                        & Power & Hours  & Days  & KWh$\cdotp$PUE & CO\textsubscript{2}e \\
            \midrule
            \multicolumn{6}{l}{\hspace*{6mm}\em OSCAR-Based ELMos}                                                           \\[0.5mm]
            Bulgarian                                       & 1183  & 515.00 & 21.45 & 962.61         & 49.09                \\
            Catalan                                         & 1118  & 199.98 & 8.33  & 353.25         & 18.02                \\
            Danish                                          & 1183  & 200.89 & 8.58  & 375.49         & 19.15                \\
            Finnish                                         & 1118  & 591.25 & 24.63 & 1044.40        & 53.26                \\
            Indonesian                                      & 1183  & 694.26 & 28.93 & 1297.67        & 66.18                \\
            \midrule\multicolumn{6}{l}{\hspace*{6mm}\em Wikipedia-Based ELMos}                                               \\[0.5mm]
            Bulgarian                                       & 1118  & 15.45  & 0.64  & 27.29          & 1.39                 \\
            Catalan                                         & 1118  & 51.08  & 2.13  & 90.22          & 4.60                 \\
            Danish                                          & 1118  & 14.56  & 0.61  & 25,72          & 1.31                 \\
            Finnish                                         & 1118  & 21.79  & 0.91  & 38.49          & 1.96                 \\
            Indonesian                                      & 1118  & 20.28  & 0.84  & 35.82          & 1.82                 \\
            \midrule
            \multicolumn{2}{@{}l}{\textsc{Total emissions}} &       &        &       & 216.78                                \\
            \bottomrule
        \end{tabular}
    }
    \caption{Average power draw (Watts), training times (in both hours and days), mean power consumption (KWh) and CO\textsubscript{2} emissions (kg) for each ELMo model trained.}
    \label{tab:carbon}
\end{table}

In our set-up we used two different machines, each one having 4 NVIDIA GeForce GTX 1080 Ti graphic cards and 128GB of RAM, the difference between the machines being that one uses a single Intel Xeon Gold 5118 processor, while the other uses two Intel Xeon E5-2630 v4 processors. One GeForce GTX 1080 Ti card is rated at around 250 W,\footnote{\url{https://www.geforce.com/hardware/desktop-gpus/geforce-gtx-1080-ti/specifications}} the Xeon Gold 5118 processor is rated at 105 W,\footnote{\url{https://ark.intel.com/content/www/us/en/ark/products/120473/intel-xeon-gold-5118-processor-16-5m-cache-2-30-ghz.html}} while one Xeon E5-2630 v4 is rated at 85 W.\footnote{\url{https://ark.intel.com/content/www/us/en/ark/products/92981/intel-xeon-processor-e5-2630-v4-25m-cache-2-20-ghz.html}} For the DRAM we can use the work of \citet{Desrochers:2016} to estimate the total power draw of 128GB of RAM at around 13W. Having this information, we can now use the formula proposed by \citet{strubell:2019} in order to compute the total power required to train one ELMo model:
\[
    p_t = \frac{1.58t(cp_{c} + p_r + gp_g)}{1000}
\]
Where $c$ and $g$ are the number of CPUs and GPUs respectively, $p_c$ is the average power draw (in Watts) from all CPU sockets, $p_r$ the average power draw from all DRAM sockets, and $p_g$ the average power draw of a single GPU. We estimate the total power consumption by adding GPU, CPU and DRAM consumptions, and then multiplying by the \emph{Power Usage Effectiveness} (PUE), which accounts for the additional energy required to support the compute infrastructure. We use a PUE coefficient of 1.58, the 2018 global average for data centers \citep{strubell:2019}. In table \ref{tab:carbon} we report the training times in both hours and days, as well as the total power draw (in Watts) of the system used to train each individual ELMo model. We use this information to compute the total power consumption of each ELMo, also reported in table \ref{tab:carbon}.

We can further estimate the CO\textsubscript{2} emissions in kilograms of each single model by multiplying the total power consumption by the average CO\textsubscript{2} emissions per kWh in France (where the models were trained). According to the RTE (Réseau de transport d'électricité / Electricity Transmission Network) the average emission per kWh were around 51g/kWh in November 2019,\footnote{\url{https://www.rte-france.com/fr/eco2mix/eco2mix-co2}} when the models were trained. Thus the total CO\textsubscript{2} emissions in kg for one single model can be computed as:
\[
    \text{CO}_{2}\text{e} = 0.051 p_t
\]
All emissions for the ELMo models are also reported in table \ref{tab:carbon}.

We do not report the power consumption or the carbon footprint of training the UDPipe 2.0 architecture, as each model took less than 4 hours to train on a machine using a single NVIDIA Tesla V100 card. Also, this machine was shared during training time, so it would be extremely difficult to accurately estimate the power consumption of these models.

Even though it would have been interesting to replicate all our experiments and computational cost estimations with state-of-the-art fine-tuning models such as BERT, XLNet, RoBERTa or ALBERT, we recall that these transformer-based architectures are extremely costly to train, as noted by the BERT authors on the official BERT GitHub repository,\footnote{\url{https://github.com/google-research/bert}} and are currently beyond the scope of our computational infrastructure. However we believe that ELMo contextualized word embeddings remain a useful model that still provide an extremely good trade-off between performance to training cost, even setting new state-of-the-art scores in parsing and POS tagging for our five chosen languages, performing even better than the multilingual mBERT model.

\section{Conclusions}

In this paper, we have explored the use of the Common-Crawl-based OSCAR corpora to train ELMo contextualized embeddings for five typologically diverse mid-resource languages. We have compared them with Wikipedia-based ELMo embeddings on two classical NLP tasks, POS tagging and parsing, using state-of-the-art neural architectures. Our goal was to explore whether the noisiness level of Common Crawl data, often invoked to criticize the use of such data, could be compensated by its larger size; for some languages, the OSCAR corpus is several orders of magnitude larger than the corresponding Wikipedia. Firstly, we found that when properly filtered, Common Crawl data is not massively noisier than Wikipedia. Secondly, we show that embeddings trained using OSCAR data consistently outperform Wikipedia-based embeddings, to the extent that they allow us to improve the state of the art in POS tagging and dependency parsing for all the 6 chosen treebanks. Thirdly, we observe that more training epochs generally results in better embeddings even when the training data is relatively small, as is the case for Wikipedia.

Our experiments show that Common-Crawl-based data such as the OSCAR corpus can be used to train high-quality contextualized embeddings, even for languages for which more standard textual resources lack volume or genre variety. This could result in better performances in a number of NLP tasks for many non highly resourced languages.

\subsection*{Acknowledgments}

We want to thank Ganesh Jawahar for his insightful comments and suggestions during the early stages of this project. This work was partly funded by the French national ANR grant BASNUM (\mbox{ANR-18-CE38-0003}), as well as by the last author's chair in the PRAIRIE institute,\footnote{\url{http://prairie-institute.fr/}} funded by the French national ANR as part of the ``Investissements d’avenir'' programme under the reference \mbox{ANR-19-P3IA-0001}. The authors are grateful to Inria Sophia Antipolis - Méditerranée ``Nef''\footnote{\url{https://wiki.inria.fr/wikis/ClustersSophia}} computation cluster for providing resources and support.

\bibliography{ELMos}
\bibliographystyle{acl_natbib}

\appendix

\section{Appendix}
\label{sec:appendix}
\subsection{Number of training steps for each checkpoint and each corpus}

\begin{table}[ht!]
    \centering\small
    \scalebox{0.96}{
        \begin{tabular}{@{}lrrrr@{}}\toprule
            Language   & 1 Epoch & 3 Epochs & 5 Epochs  & 10 Epochs        \\
            \midrule
            \multicolumn{5}{l}{\hspace*{6mm}\em Wikipedia-Based ELMos}     \\[0.5mm]
            Bulgarian  & 6,268   & 18,804   & 31,340    & 62,680           \\
            Catalan    & 20,666  & 61,998   & 103,330   & 206,660          \\
            Danish     & 5,922   & 17,766   & 29,610    & 59,220           \\
            Finnish    & 8,763   & 26,289   & 43,815    & 87,630           \\
            Indonesian & 7,891   & 23,673   & 39,455    & 78,910           \\
            \midrule\multicolumn{5}{l}{\hspace*{6mm}\em OSCAR-Based ELMos} \\[0.5mm]
            Bulgarian  & 143,169 & 429,507  & 715,845   & 1,431,690        \\
            Catalan    & 81,156  & 243,468  & 405,780   & 811,560          \\
            Danish     & 81,156  & 243,468  & 405,780   & 811,560          \\
            Finnish    & 181,230 & 543,690  & 906,150   & 1,812,300        \\
            Indonesian & 263,830 & 791,490  & 1,319,150 & 2,638,300        \\
            \bottomrule
        \end{tabular}
    }
    \caption{Number of training steps for each checkpoint, for the \elmowiki and \elmooscar of each language.}
    \label{tab:steps}
\end{table}

\end{document}